\documentclass[10pt,twocolumn,letterpaper]{article}

\usepackage[pagenumbers]{cvpr} %
\makeatletter
\@namedef{ver@everyshi.sty}{}
\makeatother
\usepackage{tikz}

\usepackage{xcolor}

\usepackage{graphicx}
\usepackage{amsmath}
\usepackage{amssymb}

\usepackage{siunitx}
\usepackage{adjustbox}
\usepackage{booktabs}
\usepackage{multirow}
\usepackage{subcaption}
\usepackage{multirow}
\usepackage{makecell}
\usepackage{pifont}%

\usepackage{etoolbox}
\robustify\bfseries

\usepackage{enumitem}

\usepackage{url}
\usepackage{cite}

\usepackage[english]{babel}

\def\MYTITLE{Low-power, Continuous Remote Behavioral Localization with Event Cameras}

\definecolor{cvprblue}{rgb}{0.21,0.49,0.74}

\usepackage[pagebackref=false,breaklinks=true,colorlinks,,citecolor=cvprblue, bookmarks=true,bookmarksnumbered=true]{hyperref}
\hypersetup{
  pdftitle={Event Penguins: \MYTITLE},
  pdfsubject={Computer Vision, Artificial Intelligence},
  pdfauthor={F. Hamann, S. Ghosh, I. Juarez-Martinez, T. Hart, A. Kacelnik, G. Gallego},
  pdfkeywords={Event Cameras, Intelligent Systems, Penguins, Remote Behavioral Monitoring}
}

\usepackage[capitalize]{cleveref}
\crefname{section}{Sec.}{Secs.}
\Crefname{section}{Section}{Sections}
\Crefname{table}{Table}{Tables}
\crefname{table}{Tab.}{Tabs.}

\newif\ifshowmainpaper %
\showmainpapertrue

\newif\ifshowsupplementary
\showsupplementarytrue

\usepackage[absolute]{textpos}

\newcommand{\unum}[1]{\underline{\num{#1}}}
\newcommand{\bnum}[1]{\bfseries #1}

\def\pol{p} %

\def\cE{\mathcal{E}} %

\def\tstart{t^{a}}
\def\tend{t^{b}}

\def\tstartAug{t^{\prime a}} %
\def\tendAug{t^{\prime b}}

\def\interval{I}
\def\framerep{\mathbf{H}} %

\newcommand{\cmark}{\ding{51}}%
\newcommand{\xmark}{\ding{55}}%

\usepackage{xcolor}
\definecolor{light-gray}{gray}{0.6}
\newcommand\gframe[1]{{\color{light-gray}\frame{#1}}}

\begin{document}

\title{\MYTITLE}

\definecolor{somegray}{gray}{0.5}
\newcommand{\darkgrayed}[1]{\textcolor{somegray}{#1}}
\begin{textblock}{11}(2.5, 0.8)  %
\begin{center}
\darkgrayed{This paper has been accepted for publication at the\\
IEEE Conference on Computer Vision and Pattern Recognition (CVPR), Seattle, 2024.
\copyright IEEE}
\end{center}
\end{textblock}

\author{Friedhelm Hamann$^{1, 5}$, Suman Ghosh$^{1}$, Ignacio Ju\'arez Mart\'inez$^{2}$,\\
Tom Hart$^{3}$, Alex Kacelnik$^{2, 5}$ and Guillermo Gallego$^{1,4, 5}$.\\
$^{1}$~Technische Universit\"at Berlin, $^{2}$~Oxford University, $^{3}$~Oxford Brookes University,\\
$^{4}$~Einstein Center for Digital Future, $^{5}$~Science of Intelligence Excellence Cluster.
}
\maketitle

\ifshowmainpaper

\begin{abstract}
   Researchers in natural science need reliable methods for quantifying animal behavior. 
   Recently, numerous computer vision methods emerged to automate the process.
   However, observing wild species at remote locations remains a challenging task due to difficult lighting conditions and constraints on power supply and data storage.
   Event cameras offer unique advantages for battery-dependent remote monitoring due to their low power consumption and high dynamic range capabilities.
   We use this novel sensor to quantify a behavior in Chinstrap penguins called ecstatic display. 
   We formulate the problem as a temporal action detection task, determining the start and end times of the behavior.
   For this purpose, we recorded a colony of breeding penguins in Antarctica for several weeks and labeled event data on 16 nests.
   The developed method consists of a generator of candidate time intervals (proposals) and a classifier of the actions within them.
   The experiments show that the event cameras' natural response to motion is effective for continuous behavior monitoring and detection, reaching a mean average precision (mAP) of 58\% (which increases to 63\% in good weather conditions).
   The results also demonstrate the robustness against various lighting conditions contained in the challenging dataset. 
   The low-power capabilities of the event camera allow it to record significantly longer than with a conventional camera.
   This work pioneers the use of event cameras for remote wildlife observation,
   opening new interdisciplinary opportunities. \url{https://tub-rip.github.io/eventpenguins/}
\end{abstract}

\section{Introduction}

Quantification of wildlife animal behavior is key to several disciplines from ecology \cite{Marks2010ecstatic} to conservation \cite{caravaggi2017review} and neuroscience \cite{Pereira20nature}.
Behavioral studies need to know when, where, and for how long a particular behavior occurs to understand it or to draw conclusions about animal relationships \cite{jouventin2017penguins}, feeding habits \cite{doniol2011optimal}, welfare \cite{dawkins2004using} or culture \cite{whiten2021burgeoning}.

\def\figmethodwidth{.48\linewidth}
\begin{figure}[t]
   \centering
   {\includegraphics[trim={8.3cm 5.8cm 11.7cm 0},clip,width=.9\linewidth]{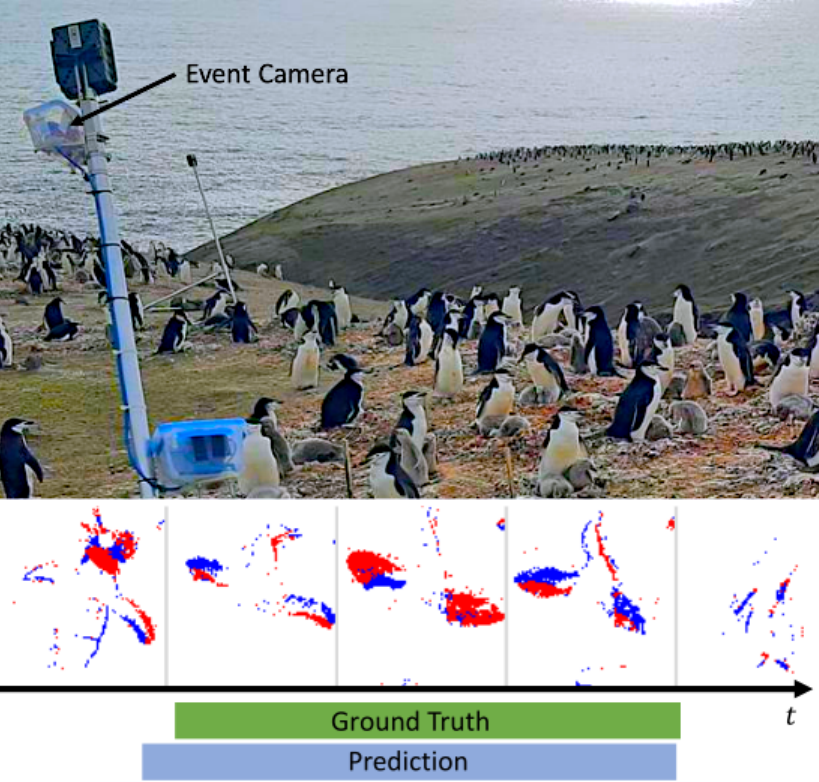}}
\caption{\emph{Remote wildlife monitoring using an event camera}. 
Top: We recorded a colony of breeding Chinstrap penguins in Antarctica using a DAVIS346 camera.
Bottom: Penguins show a behavior called ``ecstatic display'' (ED), visualized here using event data (positive events in blue, negative events in red).
We provide the dataset with labels (Ground Truth) and investigate methods to determine the start and end times of this behavior (Prediction).}
\label{fig:introduction:eyecatch}
\vspace{-1ex}
\end{figure}
Advances in computer vision have recently led to the development of various systems (data acquisition and processing) for automated quantification of animal behavior \cite{Pereira20nature,Weinstein18jae,Kay22eccv}.
It is a challenging topic because:
(i) systems for wildlife monitoring are expected to work under varying lighting and weather conditions (snow, rain, fog),
(ii) long-term observation of animals at remote places puts constraints on power usage and storage of the recording system,
(iii) extracting biologically relevant information from large amounts of data is difficult and time-consuming,
(iv) there is a lack of curated datasets with ground truth labels for developing robust machine learning systems.

We tackle the problem of quantifying the Ecstatic Display (ED) in Chinstrap penguins (\emph{Pygoscelis antarctica}) during a month of their breeding season in Antarctica.
The ED is a unique behavior where nesting penguins stand upright, point their heads upwards, beat their wings back and forth, and emit a loud call (see \cref{fig:introduction:eyecatch} and the accompanying \textbf{video}).
The reasons behind this behavior are not well understood yet \cite{jouventin2017penguins}.
Thus, we aim to better understand this behavior through its large-scale automatic detection. 
We formulate the problem as a temporal action detection (TAD) task, which consists of localizing action instances (e.g., ED in each nest) from raw data streams (usually videos).
TAD is a challenging computer vision task because it should estimate precise start and end times of the action.
Moreover, TAD for ED is challenging because:
(i) the ED length of different instances varies widely, in our data it ranges from approximately 1 to 40 s,
(ii) the birds show similar wing-flapping behaviors that are not EDs,
and are difficult to tell apart using a single snapshot (see \cref{fig:introduction:ecstatic_display}). 
Only the availability of continuous temporal information allows us to reliably differentiate instances of ED from other wing flaps. 

Current observation methods for animal behavior in the wild mostly rely on camera traps that produce a series of motion-triggered images. 
Long-term observation (weeks) of penguin behavior is provided by systems acquiring one image per minute.
Such battery-powered systems suffer from energy constraints, which are increased by the need for illuminating with IR flashes in dim-light conditions.
Neither of the above methods is suitable for analyzing fine-scale behaviors like EDs because the time resolution is insufficient or the covered observation time is too short. 

We propose using event cameras for the quantification of behaviors like the ED.
Event cameras \cite{Lichtsteiner08ssc} are novel bio-inspired sensors that output pixel-wise intensity changes instead of frames at a fixed rate.
By design, they offer several advantages, such as high speed, high dynamic range (HDR), low power consumption, and data efficiency \cite{Gallego20pami}.
Wildlife %
behavior monitoring
is a perfect scenario for event cameras and their natural property of highlighting movement.
To this end, we used a DAVIS346 camera \cite{Taverni18tcsii} to record a colony of Chinstrap penguins in Antarctica.
The annotated data consists of 24 sequences, of ten minutes each, observing 16 penguin pairs breeding in nests.
Besides providing this unique and difficult-to-acquire data, we introduce a method to detect instances of ED in two stages: 
generation of candidate pairs of start and end times of the behavior, and classification of the data within the pairs into ED or background.

The experiments show that the event data provides a distinct signal for ED quantification in terms of naturally capturing motion and being robust to challenging outdoor conditions.
The results evidence a solid performance across a variety of tests, with a mean average precision (mAP) of 58\%.
A performance comparison during the night and with snow, together with a qualitative comparison with grayscale frames, show robustness to varying illumination conditions.

Our contributions are summarized as follows:
\begin{enumerate}[noitemsep]
    \item The first-ever use of event cameras for wild animal behavior observation.
    The system for energy-efficient continuous recording overcomes the limitations of previous frame-based solutions.
    \item A method for the task of temporal action detection, consisting of proposal and classification stages.
    Both stages rely on efficient algorithms that make use of the characteristics of event data
    (\cref{sec:method}).
    \item An extensive (weeks long) event camera dataset of a Chinstrap penguin colony in Antarctica during the breeding season (\cref{sec:dataset}). 
    Twenty-four sequences of 10 minutes each are annotated with instances of ED on 16 nests, constituting a benchmark to foster research in this important wildlife monitoring field (\cref{sec:experim}).  
\end{enumerate}

To the best of our knowledge, this is the first publicly available dataset for event-based TAD, and it is also the first time for an event camera in Antarctica, 
which is a milestone demonstrating the viability of this technology for remote monitoring in extreme conditions.
Our dataset and system contribute to advancing the knowledge about aspects of penguin behavior that would have been otherwise too costly (in both time and resources) to study. 

\section{Related Work}

\begin{figure}[t]
\centering
\begin{subfigure}{0.22\linewidth}
  \centering
  {\includegraphics[trim={2cm 0cm 4cm 8mm},clip, height=1.4\linewidth]{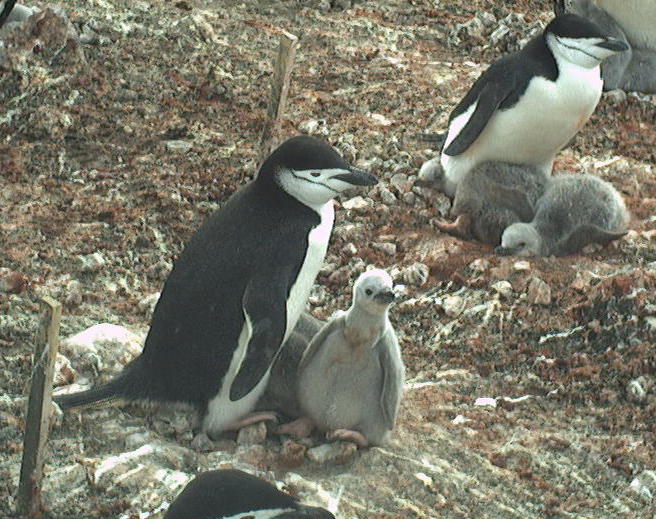}}
  \caption{Resting}
\end{subfigure}%
\begin{subfigure}{0.22\linewidth}
  \centering
  {\includegraphics[trim={2cm 2cm 3cm 1cm},clip, height=1.4\linewidth]{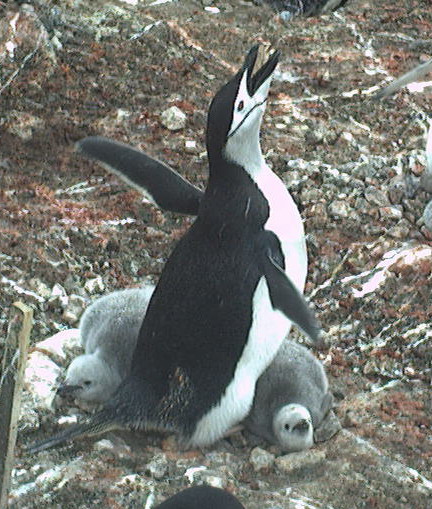}}
  \caption{ED}
\end{subfigure}\,%
\begin{subfigure}{0.22\linewidth}
  \centering
  {\includegraphics[trim={1.5cm 2cm 1.5cm 1cm},clip, height=1.4\linewidth]{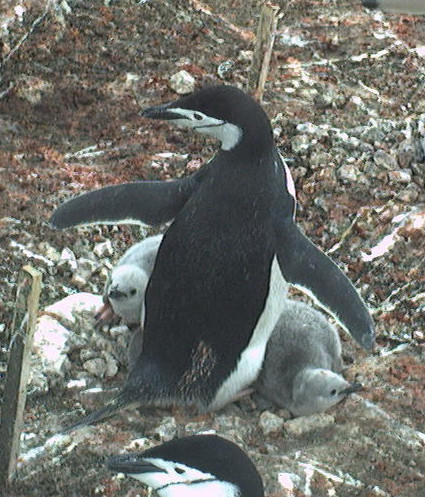}}
  \caption{Casual flap}
  \label{fig:method:overlay}
\end{subfigure}\;\;\;
\begin{subfigure}{0.22\linewidth}
  \centering
  {\includegraphics[trim={2cm 2cm 3cm 15mm},clip, height=1.4\linewidth]{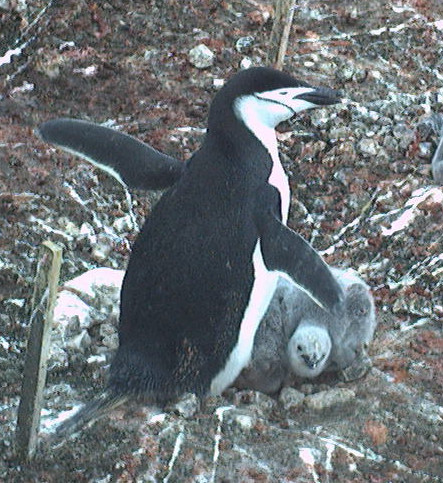}}
  \caption{ED (hard)}
\end{subfigure}
\caption{
\emph{Challenges in detecting Ecstatic Display (ED):} 
a) Penguin in rest position at its nest. 
b) Penguin showing an ED, a behavior where it points its head upwards and beats its wings while emitting a loud call. 
c) Not an ED.
d) The start of an ED. 
Comparing c) and d) shows why it is difficult to distinguish EDs from other wing flaps using single snapshots.}
\label{fig:introduction:ecstatic_display}
%\vspace{-1ex}
\end{figure}

\subsection{Event Cameras}

Event cameras are a relatively new technology. 
Since the seminal work \cite{Lichtsteiner08ssc} they have gained increasing interest due to their appealing properties, which allow them to perform well in challenging scenarios for standard cameras, such as high speed, high dynamic range (HDR), and low power consumption.
They have been studied for various applications in computer vision, %
like optical flow estimation, depth estimation, and SLAM. See \cite{Gallego20pami} for a recent survey.

Research problems studied in the context of pattern recognition mainly target classification (e.g., recognition) \cite{Orchard15fns,Amir17cvpr,Vasudevan22paa,Miao19fnbot} and detection \cite{De20arxiv,Perot20neurips}.
Due to the novelty of this imaging technology, efforts to use event cameras for pattern recognition tasks are limited by the availability of datasets. 
To this end, \cite{Orchard15fns} use a pan-tilt camera platform to convert the conventional frame-based dataset into an event dataset. 
Others provide data for gesture recognition \cite{Amir17cvpr} or sign language recognition \cite{Vasudevan22paa}.
Larger datasets are available in the context of automotive object classification and detection tasks \cite{De20arxiv,Perot20neurips, Sironi18cvpr}.
Despite recent efforts by the research community to provide datasets for various pattern recognition tasks, there is still a lack of datasets for many applications involving a variety of environmental conditions as well as data streams longer than a few seconds.
\subsection{Wildlife Observation using Computer Vision}

Automated data acquisition for wildlife monitoring currently relies on recording color (i.e., RGB) images mostly through the use of camera traps and drones.
Data is then hand-annotated \cite{Sollmann18aje}, generally aided by various computer programs.
Only recently has computer vision entered the field \cite{Weinstein18jae} with applications in counting \cite{Arteta16eccv,Kay22eccv}, species classification \cite{Beery18cvprw,Berg14cvpr,Nilsback06cvpr,Swanson15nature}, detection \cite{Bondi20wacv,Beery18eccv}, individual identification \cite{Parham17aaai,Holmberg09esr,Li19arxiv2} and tracking \cite{Kay22eccv,Bondi20wacv}.
Of all these, only \cite{Arteta16eccv} has used computer vision to count penguins,
a task performed on single static images, which is simpler than our behavior monitoring problem.
Other current approaches for automated behavior monitoring involve mounting and retrieval of tri-axial accelerometers to learn behavior classification \cite{chessa2017comparative,sutton2021fine}. 
This invasive, non-vision approach can also monitor at-sea behavior, but it incurs a much greater effort, cost, and disruption to the animals for monitoring a similar amount of individuals for an equivalent amount of time.

\subsection{Temporal Action Detection}

The goal of Temporal Action Detection (TAD) is to determine the label and time interval of every action instance in an untrimmed video.
It has been extensively studied for conventional RGB data.
Previous methods can be roughly categorized into bottom-up and top-down approaches.
Bottom-up approaches first perform frame-level classification and then merge the results into detections \cite{Yuan17cvpr,Lin21cvpr}.
Top-down approaches can have one \cite{Lin17acm,Long19cvpr} or multiple stages \cite{Xu20cvpr}, and usually follow a scheme where proposals are generated and fed to a classifier \cite{Zhao20ijcv,Lin18eccv,Lin19iccv}.
Recently, end-to-end learned methods were introduced \cite{Liu22tip, Zhang22eccv, Cheng22eccv}.
TAD for conventional RGB data is a mature computer vision task, aided by large-scale datasets, such as \cite{Jiang14crcv,Heilbron15cvpr}.

\begin{figure}[t]
    \centering
\begin{subfigure}{0.48\linewidth}
  \centering
  \includegraphics[width=\linewidth]{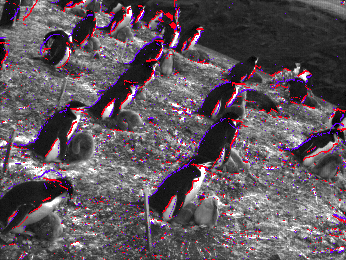}
  \caption{Events overlaid on frames.\label{fig:dataset:example_data}}
\end{subfigure}
\begin{subfigure}{0.48\linewidth}
  \centering
  \includegraphics[width=\linewidth]{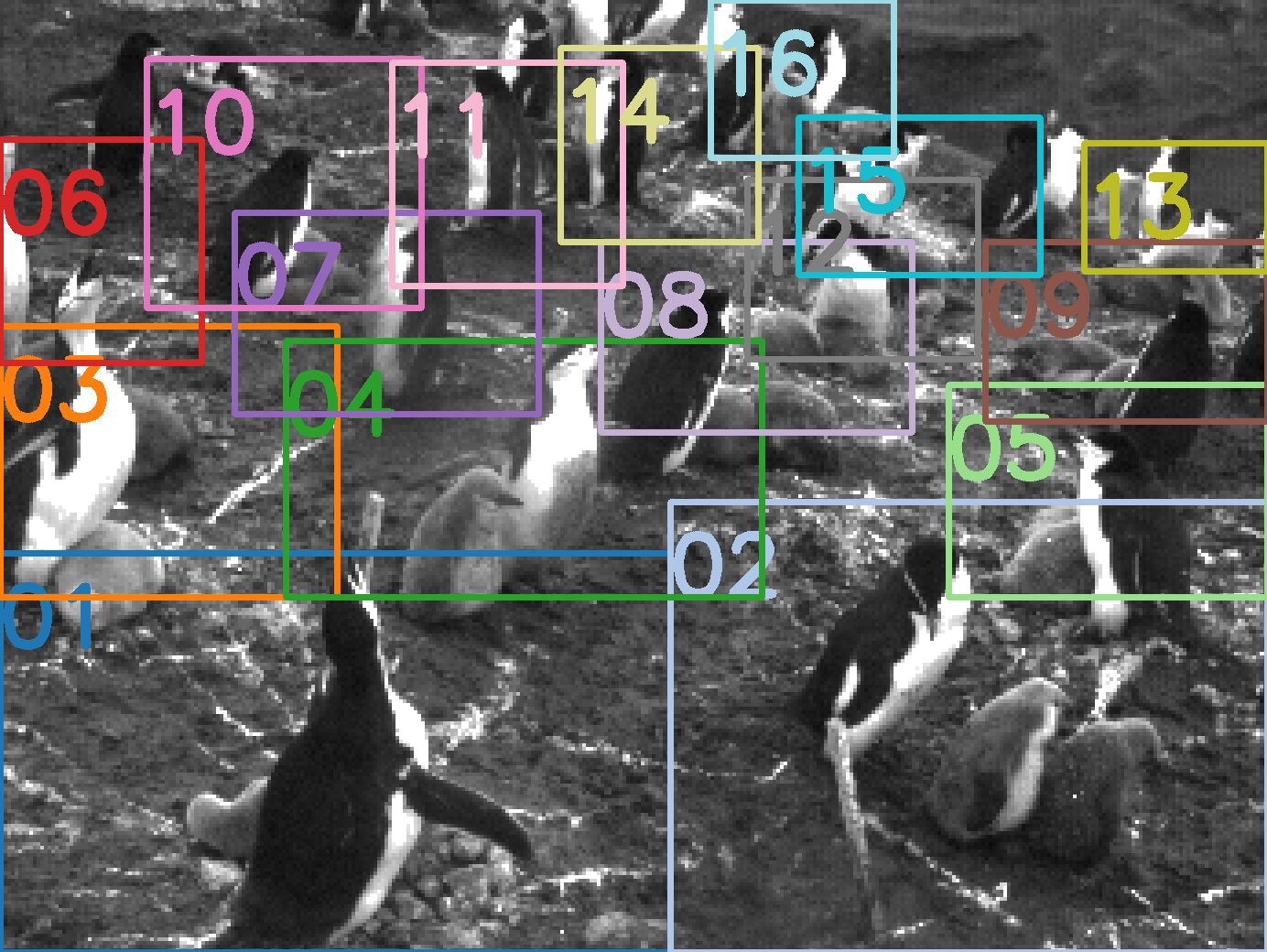}
  \caption{Selected nests.\label{fig:dataset:heatmap_annotations}}
\end{subfigure}%
    \caption{(a) Raw data acquired by the DAVIS346: 
events (in red and blue according to polarity, with \si{\micro\second} accuracy) overlaid on grayscale frames (1 every 8s).
(b) Bounding boxes around the nests.
A heat map of events over a whole sequence served as a base to hand-annotate the bounding boxes, each capturing one nest.}
    \label{fig:dataset}
\end{figure}

In the context of animal observation, \cite{Liu20bse} performed spatiotemporal action recognition to detect tail-biting behavior in pigs in a controlled indoor environment.
The work in \cite{Chen22toc} used event-based TAD for human fall detection indoors; 
the method consisted of proposal and classification stages.
For the proposals, they adopted an algorithm called temporal actioness grouping (TAG), introduced in \cite{Zhao20ijcv}, and used the event rate as the actioness score.

To the best of our knowledge, there is no publicly available dataset for animals in the wild using event cameras, a scenario with more disturbances than in indoor conditions.
With this work, we demonstrate the possibilities that event cameras offer for detailed wildlife monitoring and show how event data makes it easy to determine intervals of interest for TAD with computationally efficient methods.

\section{Temporal Action Detection from Event Data}
\label{sec:method}

This section formalizes the problem of temporal action detection (TAD) from event camera data (\cref{sec:method:problem}) and introduces our developed method (\cref{sec:method:method}).
  
\subsection{Problem Formulation}
\label{sec:method:problem}

Let $\cE = \{ e_i \}$ be an input sequence of events corresponding to a penguin's nest.
Each event $e_i = (x_i, y_i, t_i, \pol_i)$ contains the pixel coordinates $(x_i,y_i)$, a timestamp $t_i$ and the polarity $\pol_i$ (i.e., sign) of a brightness change. 
Let the pixel coordinates $(x_i,y_i)$ be already relative within the range defined by the bounding box coordinates of the nest (\cref{fig:dataset:heatmap_annotations}). 
Let the time interval when the penguin exhibits an ecstatic display (ED) be $\interval = (\tstart, \tend)$ (also called \emph{action instance}),
with start time $\tstart$ and end time $\tend$.
The collection of all ground-truth action instances, 
$\boldsymbol{\interval}_{\text{gt}} = \{ \interval_{n} \}_{n=1}^{N_{\text{gt}}}$,
is the temporal annotation set of $\cE$.

The goal of the temporal action detector is to generate predictions $\boldsymbol{\interval}_\text{pred}$ that cover $\boldsymbol{\interval}_\text{gt}$ accurately and completely. %
The inputs of the detector are the per-nest events, and only temporal boundaries need to be detected since spatial boundaries of the nests are given.
This procedure allows us to assign each ED detection reliably to a specific nest, thus enabling us to quantify the behavior of penguin couples in the nest.
This step is feasible because the penguins remain in their respective nests while breeding.

\subsection{Method}
\label{sec:method:method}

We follow a two-step approach (\cref{fig:method}). 
First, proposals are created based on an actioness score (\cref{sec:method:regionproposals}). 
Then, they are classified as ED or ``background'' by our Augmented Temporal Segment Network (ATSN) 
(\cref{sec:method:atsn}).

\subsubsection{Generation of Temporal Region Proposals}
\label{sec:method:regionproposals}
The natural response of event cameras to motion is used to efficiently produce relevant proposals. 
Letting the event rate of $\cE$ be $r(t)$ (measured in events/s), 
in the first step, an actioness score is calculated as a normalized event rate by
\begin{equation}
\tilde{r}(t) = \frac{r(t) - \min(r(t))}{\max(r(t)) - \min(r(t))}.
\label{eq:method:rate}
\end{equation}
Note that we use the robust minimum and maximum, defined by the $p$-th and $(100-p)$-th percentile, clipping $r(t)$ outside these values, and that polarity is not considered.

In the second step (ii), the intervals of high event rate (i.e., activity) are used as proposals: 
using the watershed algorithm \cite{Roerdink00fi}, 
$\tilde{r}(t)$ can be viewed as a terrain which is ``flooded'' up to a water level (threshold) $\lambda$.
The regions $\tilde{r}(t) > \lambda$ are used as proposals.
In the third step (iii), the proposals are merged to avoid short interruptions.
Starting from a proposal $\hat{\interval}_1=(\tstart_{1}, \tend_{1})$, it is iteratively merged with a subsequent one $\hat{\interval}_2=(\tstart_{2}, \tend_{2})$ if the ratio of the sum of individual durations over the duration of the merged proposal is above the merging threshold $\mu$.

Lastly, %
steps (ii) and (iii) are performed for all combinations of thresholds $\lambda \in \{0.05, 0.1, \ldots, 0.95\}$ and $\mu \in \{0.05, 0.1, ..., 0.95\}$.
The output set of interval proposals $\boldsymbol{\hat{\interval}}$ is the union of the proposals obtained from all threshold combinations.
As the scheme often produces equal proposals for close threshold levels we perform non-maximum suppression (NMS) with a threshold of 0.95.

The above method, which we call \textbf{r}obust \textbf{e}vent-rate TAG (reTAG), is an extension of the video-based TAG algorithm \cite{Zhao20ijcv}.
The main differences are: 
(i) the actioness score is given by a high-temporal resolution robustly-normalized event rate that guides the proposal search while it mitigates the influence of peak rates,
and (ii) the combination~of thresholds $(\lambda,\mu,\text{NMS})$ is adapted to the event rate signal.

\begin{figure}
    \centering
    {\includegraphics[trim={0.5cm 2.1cm 0.8cm 7.8cm},clip, width=\linewidth]{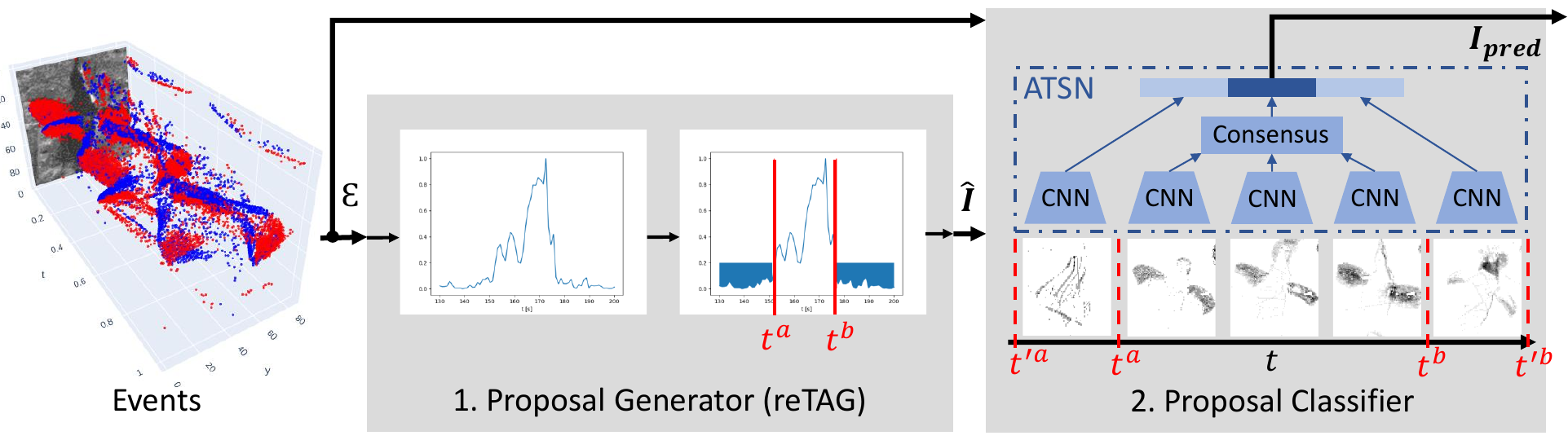}}
    \caption{\emph{Overview of the proposed method}.
    The input is the event data from one penguin nest (grayscale frame is not used and is only shown here for visualization).
    In a first step the normalized event rate is used as an actioness score to generate relevant temporal action proposals (reTAG algorithm, \cref{sec:method:regionproposals}).
    In the second step the proposals are classified (\cref{sec:method:atsn}).
    The proposals $(\tstart, \tend)$ are augmented to $(\tstartAug, \tendAug)$ and 2D event histograms are built at sampled timestamps within the augmented proposal duration. 
    These are fed to the augmented temporal segment network (ATSN). 
    }
    
    \label{fig:method}
\end{figure}

\subsubsection{Proposal Classification}
\label{sec:method:atsn}

\textbf{Overview.}  
The input to this step are the proposed time intervals $\boldsymbol{\hat{\interval}}$ and the events within them, $\cE_{\boldsymbol{\hat{\interval}}}$. 
Classifying proposals brings two challenges: 
the varying length of the proposals and the asynchronous nature of the event data.
To tackle the first challenge we leverage a common idea in action recognition and localization \cite{Wang16eccv}, sparsely sampling time snapshots from a proposal.
For the second challenge, we convert events into a tensor-like representation, which allows us to use CNN-based learning approaches.

The main component of the classification stage is a model that we term augmented temporal-spatial network (ATSN).
It works on proposals that are temporally augmented, similarly to \cite{Zhao20ijcv}, by adding ``start'' and ``end'' stages. 
The ATSN's input is the sparsely sampled tensor-like representations and its output is the proposal prediction.

\textbf{Augmentation.}   
Formally, for a given proposal $\hat{\interval} = (\tstart, \tend)$ and its duration $d = \tend - \tstart$ we define two additional intervals, the start stage $ \hat{\interval}_{\text{start}} = (\tstartAug, \tstart)$ and the end stage $\hat{\interval}_{\text{end}} = (\tend, \tendAug)$, 
where $\tstartAug = \tstart - d / W$ 
and $\tendAug = \tend + d / W$ (see \cref{fig:method}),
with $W=3$ for a 33\% augmentation width.
The augmentation of the proposals is necessary to give the classifier information about the completeness of a proposal \cite{Zhao20ijcv}: 
short proposals lying completely within an interval of an ED can be rejected based on information from the augmented intervals.
Each augmented proposal consists of three consecutive intervals: $\{\hat{\interval}_{\text{start}}, \hat{\interval}, \hat{\interval}_{\text{end}} \}$.

\textbf{Sparse Sampling.} 
For each augmented proposal, we uniformly sample $N_{\hat{\interval}}=3$ timestamps in $\hat{\interval}$ and $N_{\hat{\interval}_{\text{start}}}=N_{\hat{\interval}_{\text{end}}}=1$ timestamps in $\hat{\interval}_{\text{start}}$ and $\hat{\interval}_{\text{end}}$. 
This step is independent of the actual duration of the proposal. 
Hence, the time between the timestamps can vary for different proposals.

\textbf{Tensor-like representation.} 
We convert events into tensor-like representations for compatibility with CNN networks.
At each sampled timestamp $t_i$, a histogram \cite{Maqueda18cvpr} or a time map \cite{Lagorce17pami} of events $\framerep$ is computed using a window %
$[t_i - \Delta t/2,  t_i + \Delta t/2]$.
These snapshots are fed to the ATSN. 

\textbf{ATSN.} 
The ATSN receives the $N_{\hat{\interval}}+N_{\hat{\interval}_{\text{start}}}+N_{\hat{\interval}_{\text{end}}}=5$ snapshots, encodes them with a backbone ResNet \cite{He16cvpr} and combines the resulting feature vectors (\cref{fig:method}).
The features from the central interval $\hat{\interval}$ are aggregated via a consensus function (e.g., the mean), 
followed by a concatenation of the feature vectors corresponding to $\{\hat{\interval}_{\text{start}}, \hat{\interval}, \hat{\interval}_{\text{end}} \}$.
The resulting vector is the input to the fully connected classifier.
The final step is a non-maximum suppression (NMS) with a temporal Intersection over union (tIoU) threshold of 0.6.

\section{Dataset}
\label{sec:dataset}

\subsection{Data Collection}

\begin{table}[t!]
\centering
\adjustbox{max width=\columnwidth}{%
\setlength{\tabcolsep}{4pt}
\begin{tabular}{l*{6}{S[detect-weight,detect-mode]}}
\toprule
 Day                   & \text{\#seqs} &\text{\#night} & \text{\#precipi-} & \text{Mean rate} & \text{Max rate} & \text{Avg. \#ED} \\
                    &  &  & \text{tation} & \text{[$\times 10^3$ev/s]} & \text{[$\times 10^3$ev/s]} & \text{per seq.}\\
\midrule 
Jan 5th             &       \text{1}  &        \text{0}  &     \text{0} & 4.819276667 & 98.52 & 2 \\
Jan 6th             &       \text{2} &          \text{1}  &      \text{0} & 22.684724165 & 98.52 &  32 \\
Jan 7th             &       \text{3}  &        \text{1}  &     \text{1} & 10.846929999 & 154.14 & 31 \\
Jan 9th             &       \text{1}  &        \text{0}  &     \text{0} & 11.290095 & 28.17 & 28\\
Jan 11th            &      \text{1}   &       \text{0}   &    \text{0}  & 9.730775 & 135.72 & 9\\
Jan 12th             &      \text{5}   &       \text{1}   &    \text{3}  & 9.0148696668 & 390.75 & 27 \\
Jan 13th             &      \text{4}   &       \text{1}   &    \text{1}  & 17.2258116675 & 203.64 & 38.25  \\
Jan 14th            &      \text{2}   &       \text{1}   &    \text{0}  & 16.8636583315 & 154.86 & 4.5\\
Jan 15th             &      \text{3}   &       \text{0}  &     \text{1} & 6.397052221 & 121.05 & 14.33\\
Jan 17th            &      \text{1}   &       \text{0}  &     \text{0} & 5.75979 & 121.83  & 29 \\
Jan 18th             &      \text{1}   &       \text{1} &      \text{0} & 1.85875 & 4.05 & 0 \\
\textbf{All}       &      \text{24}  &        \text{6}  &       \text{6} & 11.23789 & 390.75 & 23.54166667\\
\bottomrule
\end{tabular}
}
\caption{Statistics of the annotated data.}
\label{tab:data:dataset}
\end{table}

Data was collected during a scientific expedition to Deception Island, Antarctica. We recorded Chinstrap Penguins nesting at the Vapour Col/Punta Descubierta colony from Jan 5th to 30th, 2022. 
This covered most of the chick guarding stage and the early crèche period of the breeding season. 

Available event cameras vary in resolution and provided features \cite{Gallego20pami}. 
We chose a DAVIS346 \cite{Taverni18tcsii}, which can simultaneously record aligned events and grayscale frames (on the same array of 346$\times$260 px). 
The event camera was connected to a Raspberry Pi CPU and a hard drive. 
To waterproof these we modified a couple of Polypropylene (PP) containers by fitting a Nikon L37 46mm glass UV filter and sealing the openings with silicone (\cref{fig:introduction:eyecatch}). 
The camera was powered using a lithium-ion portable battery that needed to be changed daily, which resulted in data gaps on days when fieldwork could not be carried out due to inclement weather. 

We used the in-hardware event noise reduction filter of the DAVIS and a low contrast sensitivity setting to limit the number of events produced and therefore enable long-term monitoring. %
We set the DAVIS to record 1 grayscale frame every 8 seconds. 
In total, we acquired 238 hours of data, consisting of events and frames from the DAVIS (\cref{fig:dataset:example_data}).
Twenty-four 10-minute sequences were selected for annotation of ED, and their statistics are summarized in \cref{tab:data:dataset}.

\subsection{Data Annotation}

We labeled 16 %
nests in all sequences using hand-annotated bounding boxes (\cref{fig:dataset:heatmap_annotations}). 
Behaviors displayed outside nests or far away back were ignored. 
The label ID refers to the nest and not the penguin, as one nest is guarded in turns by two indistinguishable individuals. 
The camera shifted four times during deployment; we moved the bounding boxes accordingly for constant nest ID throughout. 
This is important for within-study consistency and to match results to other samples like blood or GPS tracks.

\begin{figure}[t]
    \centering
    {\includegraphics[trim={0cm 0.5cm 0cm 1.3cm},clip,width=0.9\linewidth]{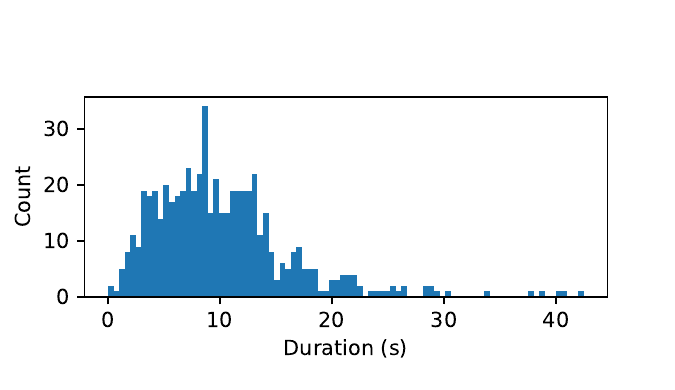}}
    \caption{Histogram of the duration of the annotated EDs. %
    Most EDs cannot be captured with previous long-term observation systems (1~frame/min).
    Events allow for continuous monitoring.
    } 
    \label{fig:dataset:ed_hist}
\end{figure}

We then annotated the occurrence of ED behavior in twenty-four 10-minute tracks (\cref{tab:data:dataset}). 
These were selected so that all hours of the day and all days in the study period were covered. 
An ED behavior was considered to last from the moment a penguin lifts its wings before flapping until the moment it stops and the wings are lowered, with a time resolution of 30 Hz.
To keep labeling consistent, all annotations were carried out by a single penguin expert and verified by two other researchers. 
\Cref{fig:dataset:ed_hist} shows a histogram of the duration of the annotations.
As \cref{fig:dataset:ed_hist} and \cref{tab:data:dataset} show, the number and duration of ED vary widely among the sequences, which contributes to making this a challenging detection problem.

\section{Experiments}
\label{sec:experim}

Several tests are carried out to assess the performance of the proposed method on the newly introduced dataset. 
This section presents the experimental settings (\cref{sec:exp:settings}) as well as the experiments for the proposal generator (\cref{sec:exp:proposals}) and the whole pipeline (\cref{sec:exp:full}). 
We also evaluate the HDR capabilities of the system (\cref{sec:exp:hdr}), 
present sensitivity studies (\cref{sec:exp:ablation}), 
report the power consumption (\Cref{sec:powerconsumption})
and discuss the limitations of our work (\cref{sec:limitations}).

\subsection{Experimental Settings}
\label{sec:exp:settings}

\textbf{Implementation Details.} 
We use a fixed data split into a training, validation, and test set (70\%:10\%:20\%).
All metrics are reported on the test data, never seen during training or validation. 
The test set contains five of the 24 sequences, including one night and one precipitation sequence (see supplementary for further details).
The ATSN is trained on a dataset consisting of all proposals generated by the first stage (\cref{fig:method}).
Proposals with an IoU $> 0.7$ are labeled as positive samples, the rest as negative.

The ATSN is trained using Stochastic Gradient Descent with a momentum of 0.9. 
The batch size is 128, and the initial learning rate is 0.001. 
The backbone of the classifier is a ResNet18, which is initialized with weights pre-trained on ImageNet~\cite{Deng09cvpr} and afterwards on the per-sample classification task. %
We used a weighted cross-entropy loss.
The bounding boxes of the penguin nests have different sizes. 
Therefore, the input event representations (histograms or time maps) are resized to 224$\times$224 px to be passed as input to the CNN.
To allow reusing the ImageNet weights, the histograms are replicated to a 3-channel input.
Experiments are conducted on hosts with an Nvidia Tesla V100S GPU and an Intel Xeon 4215R CPU. 
The network is implemented in PyTorch 1.13.0.

\begin{table}[t!]
\centering
\adjustbox{max width=\columnwidth}{%
\setlength{\tabcolsep}{5pt}
\begin{tabular}{l*{3}{S[table-format=1.2,detect-weight,detect-mode]}} %
\toprule
Proposal Method           & \text{Top 20}  & \text{Top 30}  & \text{Top 50} \\ 
\midrule 
Sliding Window            & 0.059102       & 0.066194        & 0.078014    \\
Watershed                 & \unum{0.264775}       & 0.269504        & 0.269504     \\
event TAG \cite{Chen22toc} & 0.241135        & \unum{0.330969}        & \unum{0.494090}       \\
reTAG (Ours)              & \bnum{0.430260}       &  \bnum{0.529551}       &  \bnum{0.661939}      \\
\bottomrule
\end{tabular}
}
\caption{The average recall (AR), in percentage (\%), for different proposal methods at the same number of proposals.}
\label{tab:exp:proposals}
\end{table}

\begin{figure}[t]
    \centering
    {\includegraphics[trim={.6cm 1.62cm 1.1cm 2.7cm},clip,width=.9\linewidth]{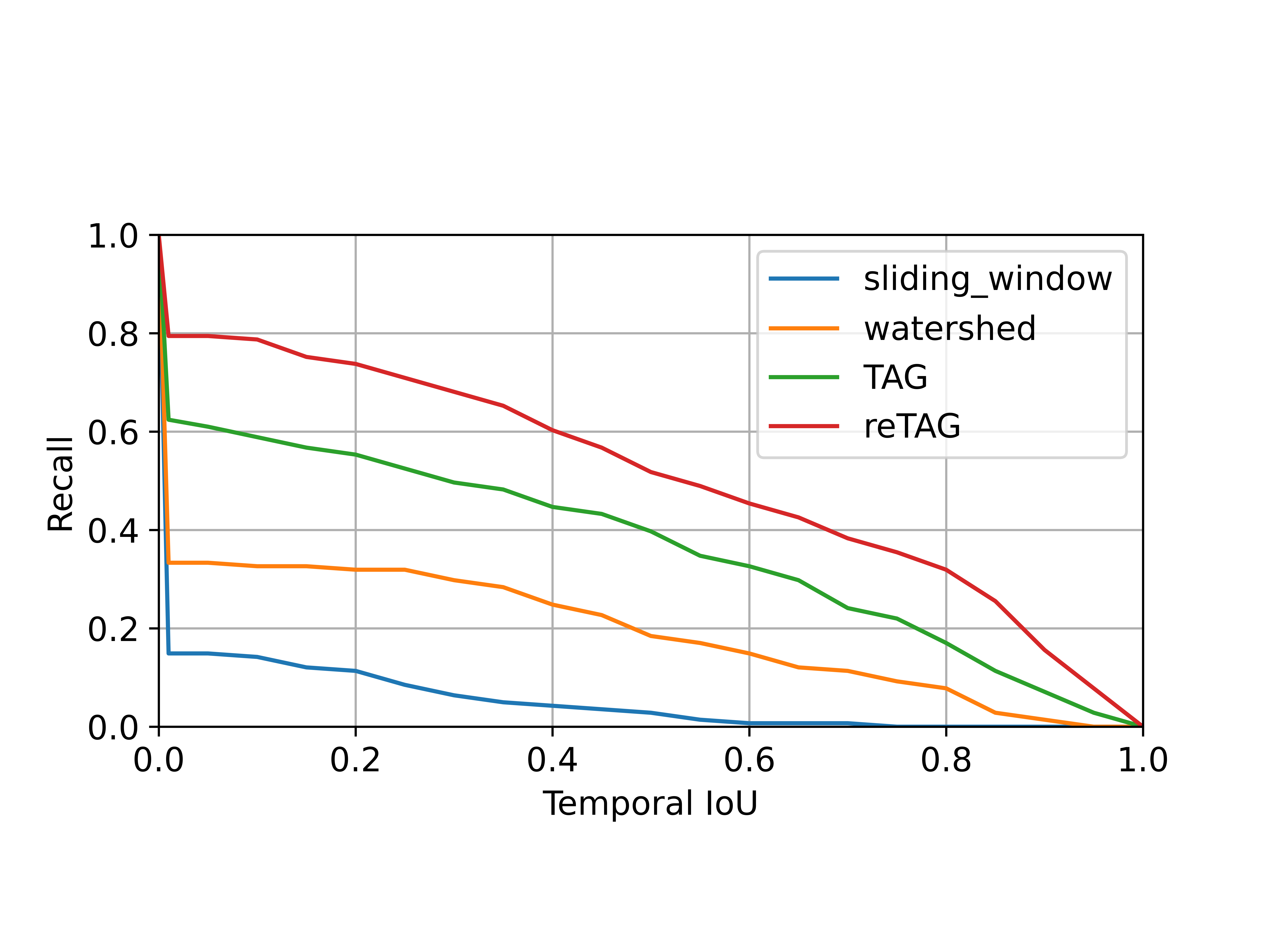}}
    \caption{The Recall over IOU rates for the top-50 proposals per recording and nest.}
    \label{fig:exp:proposal_recall}
    \vspace{-2ex}
\end{figure}

Incorporating domain knowledge, we omit intervals shorter than 2s, which account for only 2.8\% of the annotations. 
This also reduces the processing time since otherwise there are many noisy proposals.

\textbf{Evaluation Metrics.} 
We perform independent experiments for the proposal stage and the whole two-stage pipeline and evaluate using standard metrics.
For the proposal stage, we report Average Recall (AR) for $\text{tIoU} = \{0.1, 0.3, 0.5, 0.7\}$ using the best $N_p$ proposals per recording and penguin nest, with $N_p \in \{ 20, 30, 50 \}$. 
For the whole pipeline, we report temporal mean average precision (mAP) at the same tIoU values as for the AR metric.

\subsection{Evaluation of the Proposal Generator}
\label{sec:exp:proposals}

We compare our reTAG method to a re-implementation of the proposal method in \cite{Chen22toc} (since it is not publicly available) and two additional methods.
The first additional method is a bare ``watershed'' algorithm corresponding to steps (i) and (ii) in \cref{sec:method:regionproposals}, without an additional merging step or several threshold values. 
The threshold for the watershed algorithm is set to $\lambda = 0.2$, found by fitting to the training and validation sets.
The second method is a sliding-window approach \cite{Shou16cvpr} that builds proposals using all combinations of a set of start times (with stride 0.1s) and a set of window widths. 
As the range of durations varies greatly, we use a geometric progression to sample 30 window widths between 2 and 40s. 
The event rate \eqref{eq:method:rate} is computed with 33ms bins (30Hz equivalent) for all methods requiring it.

\Cref{tab:exp:proposals} shows the AR results of the different methods and \cref{fig:exp:proposal_recall} depicts the corresponding recall rates. 
Both TAG-based methods outperform the simpler methods ``watershed'' and ``sliding window''. 
A direct comparison of our reTAG with the second TAG-based method shows that adding a robust minimum and maximum (percentiles $p=1$\%) leads to significant performance gains: more than 25\% for all considered numbers of proposals. %
The result indicates that peaks in the event rate severely affect the algorithm and removing outliers boosts performance.

\subsection{Evaluation of the Full Pipeline}
\label{sec:exp:full}
\begin{table}[t!]
\centering
\adjustbox{max width=\columnwidth}{
\setlength{\tabcolsep}{5pt}

\begin{tabular}{l*{5}{S[table-format=1.2,detect-weight,detect-mode]}}
\toprule
Method                  & \text{0.1} & \text{0.3} & \text{0.5} & \text{0.7} & \text{Average} \\
\midrule 
Bottom-up (w/o MF)               &  0.5540170398249509 & 0.48322965310403504 & 0.29799806854201116 & 0.19765858366306882 & 0.3832258362835165 \\
Bottom-up  (w/ MF)             & 0.5831919421925574 & 0.5435960785264372 & 0.3922602843507158 & 0.27772623909852784 & 0.4491936360420596 \\
R3D + ActionFormer &\bnum{0.7193} & \bnum{0.6486} &       0.5215 &       0.3156   &       0.5513 \\
Ours + histogram   &      0.6279  &       0.6136  & \unum{0.564} & \bnum{0.4457}  & \unum{0.5628} \\
Ours + time-map    &\unum{0.6595} & \unum{0.6379} & \bnum{0.581} & \unum{0.4251}  & \bnum{0.5759} \\
\midrule
\emph{Perfect Classifier}     & 0.9978 &  0.9716   &   0.9362 &   0.8298 &    0.93385  
\\
\bottomrule
\end{tabular}
}
\caption{\emph{Full pipeline evaluation}. 
Mean Average Precision at several IoU levels (mAP@IoU).
Best in bold. Runner-up underlined.}
\label{tab:exp:whole-pipeline}
\end{table}

To evaluate the performance of the whole two-step approach in \cref{fig:method}, we test various settings and compare the results against several baselines. 
We adapt the state-of-the-art frame-based method ActionFormer \cite{Zhang22eccv} to be used with event data and also compare against a self-designed bottom-up approach. 
Lastly, we provide results for a ``perfect classifier'' partially using ground truth information to provide an upper performance bound.

\textbf{Top-down Approach (Ours).}
Since the method in \cref{sec:method} works by first analyzing a large time interval with an actioness score and then processing finer time intervals with a classifier, it is also referred to as ``top-down''.
The method is tested with two different tensor-like representations using the events in a window of $\Delta t = 1s$.
The first representation is a 2D histogram, counting the number of events at each pixel position, 
and the second is an exponentially decaying time map \cite{Lagorce17pami} with a time constant of $\tau=0.2$s.

\textbf{Event-based ActionFormer baseline.}
We adapt ActionFormer \cite{Zhang22eccv} for use with event data.
The latter works on offline-extracted video features specific to common datasets like THUMOS14 \cite{Jiang14crcv} and ActivityNet 1.3 \cite{Heilbron15cvpr}, preventing direct adoption to custom data.
Therefore we extract features in our dataset as follows:
(i) batches of events are converted to spatiotemporal voxel grids of shape $(C,H,W)$ (height $H$, width $W$, and $C$ time bins).
Voxels are created at equidistant timesteps of 0.5s using events from a 5s time window;
(ii) each voxel grid is encoded into a 1D feature vector of size 512 using a 3D-CNN (R3D \cite{Tran18cvpr}).
The resulting sequence of feature vectors is the input to the model, which outputs the set of action instances.

\textbf{Bottom-up Approach.} 
In addition to our top-down approach of \cref{sec:method}, we build a bottom-up approach for comparison. 
The event stream $\cE$ is converted into a sequence of synchronous frames (snapshots) similar to a video. 
Afterwards, per-snapshot classification is performed using a conventional CNN (e.g., ResNet). 
Lastly, the classification results are post-processed using morphological operators along the temporal axis. 
Regions of consecutive positive results are extracted as ED predictions.

Specifically, tensor-like representations $\{\framerep_1, \ldots, \framerep_k\}$ are created by counting events in batches of $\Delta t = 5$s and with a stride of 33ms (30 Hz).
The synchronous classification predictions are then used for finding the temporal boundaries.
This method is termed ``Bottom-up (w/o MF)''. 
If we further apply a 1D closing morphological filter (MF) (with a kernel size of 15) we can join sparse predicted intervals;
a method referred to as ``Bottom-up (w/ MF).''

\textbf{Perfect Classifier.} 
This model follows a top-down approach using proposals provided by the reTAG generator. 
However, the classifier utilizes ground-truth information and classifies proposals as true if the tIoU is bigger than the respective metric value (0.1, 0.3, 0.5, 0.7). 
The classification is followed by an NMS. 
The values provide an upper bound for the classifier performance with given proposals.

\textbf{Results.} 
\Cref{tab:exp:whole-pipeline} reports the results of the abovementioned approaches. %
Our method using time maps (and the ResNet18 backbone) performs best (mAP$=58$\%), closely followed by the variant that uses histograms (mAP$=56$\%).

The top-down method outperforms the bottom-up method by 23\%. %
We found $\Delta t = 5$s works best for the bottom-up approach. 
While it %
may seem long, it is short for the simple (per-sample) classification task, indicating that the bottom-up method does not aggregate temporal information meaningfully (simply increasing $\Delta t$ does not improve performance).
The ATSN has information from multiple sampled timestamps and therefore yields better results.

Our method outperforms ActionFormer by 3\% on average (\cref{tab:exp:whole-pipeline}).
This could be due to the amount of labeled data.
State-of-the-art frame-based methods in action detection (e.g., ActionFormer) rely on large amounts of data (e.g., ActivityNet $\approx$648h).
However, this is often unfeasible for use cases in conservation.
Our approach relies on a non-learned first stage and a second stage with only 11M parameters (compared to 27M parameters of ActionFormer).
Methods needing less data are more prone to be adopted by researchers in biology. 
Finally, the results show that our method performs better for higher IoU thresholds, yielding more accurate interval boundaries.
Per-nest results are given in the supplementary.

\subsection{High Dynamic Range Experiment}
\label{sec:exp:hdr}

The dataset contains samples from various lighting and weather conditions,
which we use to assess the robustness of our method to impaired vision.
As frame data is not available at a sufficient rate, it is not possible to directly compare a frame-based method to our event-based method.
However, we perform two examinations: 
a qualitative comparison with the grayscale frames of the DAVIS 
and a quantitative comparison of our method for different conditions.

\global\long\def\figWidth{0.31\linewidth}
\begin{figure}[t]
	\centering
    {\scriptsize
    \setlength{\tabcolsep}{1pt}
	\begin{tabular}{
	>{\centering\arraybackslash}m{0.25cm} 
	>{\centering\arraybackslash}m{\figWidth} 
    >{\centering\arraybackslash}m{\figWidth} 
	>{\centering\arraybackslash}m{\figWidth}}
 
        \rotatebox{90}{\makecell{Frame}}
		&{\includegraphics[width=\linewidth]{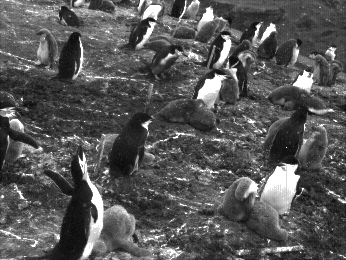}}
        &{\includegraphics[width=\linewidth]{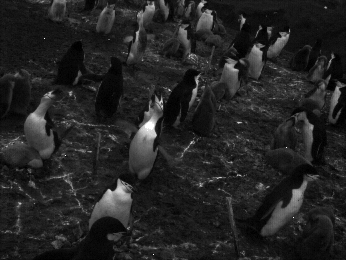}}
        &{\includegraphics[width=\linewidth]{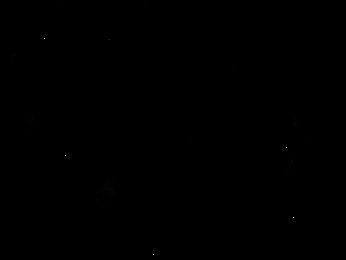}}\\
        
		\rotatebox{90}{\makecell{Events}}
        &\gframe{\includegraphics[width=\linewidth]{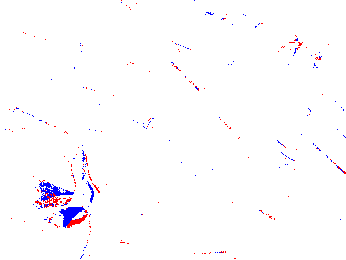}}
        &\gframe{\includegraphics[width=\linewidth]{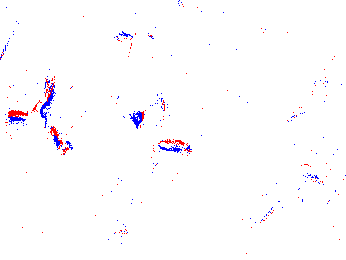}}
        &\gframe{\includegraphics[width=\linewidth]{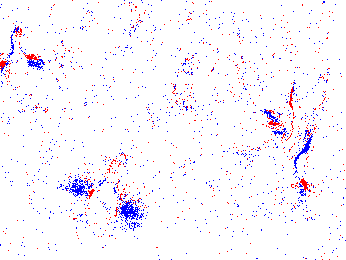}}\\
  
		& \textbf{(a)} Precipitation (snow)
        & \textbf{(b)} Dim light
		& \textbf{(c)} Night
	\end{tabular}
	}
	\caption{Examples of EDs on events and grayscale images for different lighting conditions.}
	\label{fig:exp:hdr}
\end{figure}
\begin{table}[t!]
\centering
\adjustbox{max width=\columnwidth}{%
\setlength{\tabcolsep}{5pt}
\begin{tabular}{l*{5}{S[table-format=1.2,detect-weight,detect-mode]}} %
\toprule
Category          & \text{0.1} & \text{0.3} & \text{0.5} & \text{0.7} & \text{Average} \\
\midrule 
Undisturbed       &  0.698435 & 0.687811 & 0.642922 & 0.484685 & \bnum{0.628463} \\
Night             &  0.715439 & 0.694291 & 0.631375 & 0.435573 & 0.619169 \\
Precipitation     &  0.488821 & 0.410357 & 0.338873 & 0.251714 & 0.372441 \\
\bottomrule
\end{tabular}
}
\caption{\emph{Robustness to visual conditions}.
Mean Average Precision at several IoU levels (mAP@IoU).
Undisturbed means day time with good lighting and no precipitation.
}
\label{tab:exp:sensitivity_rain_night}
\end{table}

\Cref{fig:exp:hdr} shows a comparison of frame data and event data at different visual conditions. 
The event camera signal degrades in low-light conditions and becomes significantly noisier. %
In comparison, the grayscale frame does not capture any information at the lowest lighting level.
\Cref{tab:exp:sensitivity_rain_night} shows the results on different subsets of the test set.
The category ``Undisturbed'' shows results for the three sequences at full daylight without precipitation
(the mAP increases to 63\%), 
``Night'' is the result of the night sequence and ``Precipitation'' is the result of the sequence containing snow.
In comparison to the undisturbed case, the results show a comparable performance in night conditions (mAP$=62$\%) and a decrease in the snow sequence.
This is remarkable, as it is apparent from \cref{fig:exp:hdr} that grayscale frames would be unusable (i.e., not just a 1\% drop) in night sequences. 
In summary, both examinations demonstrate the advantages of event cameras in tackling difficult illumination scenarios 
and the system's robustness (novel sensor and algorithm) to natural changes in visual conditions.

\subsection{Sensitivity and Ablation Studies}
\label{sec:exp:ablation}

We assess the sensitivity of our method to multiple design choices:
the amount of data fed to the classification network (\cref{tab:exp:sensitivity_augmentation}) and
the size of the augmentation width (\cref{tab:exp:sensitivity_augment_width}).
Additional sensitivities to the choice of network backbone 
and input representation are given in the supplementary. %

\Cref{tab:exp:sensitivity_augmentation} reports results for varying numbers of timestamp samples $N_{\hat{\interval}}$ in the ATSN. 
With increasing $N_{\hat{\interval}}$ we also increase $N_{\hat{\interval}_{\text{aug}}}$. 
The augmentation duration is fixed at 33\% of the central interval duration $d$.
The sensitivity analysis shows an increase in the performance with a higher number of samples up to $N_{\hat{\interval}} = 7$, which indicates that increasing the number of samples beyond that is ineffective.

Additionally, the first row of \cref{tab:exp:sensitivity_augmentation} reports the results for a raw TSN network without augmentation. 
Omitting augmentation severely limits the performance, thus highlighting the importance of augmentation, 
which provides information beyond the boundaries of the proposal and therefore allows the classifier to judge its ``completeness''.
\Cref{tab:exp:sensitivity_augment_width} lists the mAP for different augmentation widths $W$.
The results indicate an augmentation of 33\% as optimal.

\Cref{tab:suppl:pretraining} shows the strong influence of pretraining.

\begin{table}[t]
\centering
\adjustbox{max width=0.85\columnwidth}{%
\setlength{\tabcolsep}{5pt}
\begin{tabular}{lll*{5}{S[table-format=1.2,detect-weight,detect-mode]}} %
\toprule
$N_{\hat{\interval}}$ & $N_{\hat{\interval}_{\text{aug}}}$  &  \text{Augmented} & \text{0.1} & \text{0.3} & \text{0.5} & \text{0.7} & \text{Average} \\
\midrule 
 3       &    0    & \xmark   &  0.2853 & 0.2159 & 0.1456 & 0.04964 & 0.1741  \\
 1       &    1    & \cmark   & 0.5303 & 0.4699 & 0.395 & 0.1896 & 0.3962 \\
 3       &    1    & \cmark  & 0.6352 & 0.589 & 0.4813 & 0.3072 & 0.5032 \\
 5       &    2    & \cmark    & 0.6338 & 0.6065 & 0.5441 & \bnum{0.4318} & 0.5541 \\
 7       &    3    & \cmark \;(\cref{tab:exp:whole-pipeline})   & \bnum{0.6595} & \bnum{0.6379} & \bnum{0.581} & \unum{0.4251}  & \bnum{0.5759} \\
 9   &     3 & \cmark   & \unum{0.6534} &  \unum{0.6161} & \unum{0.5489} & 0.4039 & \unum{0.5556} \\
\bottomrule
\end{tabular}
}
\caption{Ablation and sensitivity study for the number of samples in the augmentation of the classification network. 
Mean Average Precision at several IoU levels (mAP@IoU).
}
\label{tab:exp:sensitivity_augmentation}
\end{table}

\begin{table}[t!]
\centering
\adjustbox{max width=0.85\columnwidth}{%
\setlength{\tabcolsep}{5pt}
\begin{tabular}{ll*{5}{S[table-format=1.2,detect-weight,detect-mode]}} %
\toprule
$W$ & \text{Aug.~width}   & \text{0.1} & \text{0.3} & \text{0.5} & \text{0.7} & \text{Average} \\
\midrule 
 5 & 20\%  & 0.6231 & 0.5766 & 0.5202 & \unum{0.3719} & 0.5229 \\
 3 & 33\% (\cref{tab:exp:whole-pipeline})  & \bnum{0.6595} & \bnum{0.6379} & \bnum{0.581} & \bnum{0.4251}  & \bnum{0.5759} \\
 2 & 50\%  & \unum{0.6461} & \unum{0.6086} & \unum{0.5411} & 0.3692 & \unum{0.5413} \\
\bottomrule
\end{tabular}
}
\caption{Sensitivity of the system (Ours + histogram) with respect to the augmentation width $W$.
Mean average precision at several IoU levels (mAP@IoU).}
\vspace{-1ex}
\label{tab:exp:sensitivity_augment_width}
\end{table}

\begin{table}[t!]
\centering
\adjustbox{max width=0.9\columnwidth}{%
\setlength{\tabcolsep}{5pt}
\begin{tabular}{l*{5}{S[table-format=1.2,detect-weight,detect-mode]}} %
\toprule
Pretraining strategy       & \text{0.1} & \text{0.3} & \text{0.5} & \text{0.7} & \text{Average} \\
\midrule 
None                    & 0.4526 & 0.425 & 0.3549 & 0.2586 & 0.3728    \\
Imagenet                & 0.5213 & 0.499 & 0.4527 & 0.3394 & 0.4531 \\
Imagenet + Time-map   & \bnum{0.6595} & \bnum{0.6379} & \bnum{0.581} & \bnum{0.4251}  & \bnum{0.5759} \\
\bottomrule
\end{tabular}
}
\caption{\emph{Sensitivity to pretraining of the backbone}.
Mean Average Precision at several IoU levels (mAP@IoU).
\label{tab:suppl:pretraining}
\vspace{-1ex}
}
\end{table}

\subsection{Power Consumption}
\label{sec:powerconsumption}
The DAVIS346 power usage is
0.7W in idle mode and 0.83W while recording. %
Data generation and streaming take up a small portion of the power spent by the overall device (chip, board, USB, etc.).
In comparison, a frame-based camera (Basler acA1300-200u) draws 2.28W in idle mode and 2.44W while recording.
The event camera allowed us to record significantly longer than the frame-based camera. 

\subsection{Limitations}
\label{sec:limitations}

Although our method is robust in different lighting conditions, it's not possible to record during heavy weather.
Strong wind leads to setup vibrations and noisy events.

Our detection method works under the assumption that a high event rate is linked to activity of interest.
The results show that the assumption is reasonable for observing nesting penguins.
However, it limits the generalization %
to problems where the assumption holds.

Our method relies on user-supplied regions of interest (e.g., coarse bounding boxes).
The assumption is that EDs detected in one region can be associated with a specific nest.
This has been carefully considered in the provided dataset.
Application to unseen data requires the user to select new bounding boxes with penguins not relocating between them, 
although some overlap is allowed (\cref{fig:dataset:heatmap_annotations}). %

\section{Conclusion}

We have presented the first approach for wildlife monitoring using an event camera.
The choice of sensor aims to drastically increase the ability to record continuously during long periods when relying on battery-based camera systems.
We have introduced a unique dataset of breeding Chinstrap penguins in Antarctica acquired with a DAVIS camera,
and have used it to quantify a penguin behavior called ``ecstatic display'' (ED).
The problem has been formulated as a temporal action detection task, inferring instances of the start and end times of ED.
Our two-stage detector consists of a lightweight generator of time-interval proposals and a subsequent classifier.
It outperforms comparison methods and is robust against environmental changes.
The results indicate that the event camera is a fit sensor for this problem, as it naturally captures motion and therefore simplifies the task, especially in difficult lighting conditions.

\section*{Acknowledgements}
We thank Prof.~Josabel Belliure for her help with camera deployment.
The research was funded by Deutsche Forschungsgemeinschaft (DFG, German Research Foundation)
under Germany’s Excellence Strategy – EXC 2002/1 ``Science of Intelligence'' – project number 390523135.

\fi %

\ifshowsupplementary

\section*{Supplementary Material}
\setcounter{section}{7}

\Cref{sec:suppl:bio_background} provides more information on the biological motivation of our project.
 \Cref{sec:suppl:results_per_nest} shows per nest results of our method.
In \cref{sec:suppl:runtime}, we report the number of proposals and training run time. 
\Cref{sec:suppl:dataset_details} gives more details about data acquisition, filtering, and split.
Additionally, \cref{sec:supp:sensitivity_backbone} provides several sensitivity studies.
Lastly, \cref{sec:suppl:naive_baseline} shows results for a naive baseline for the random classifier.

\subsection{Biological Motivation and Impact}
\label{sec:suppl:bio_background}
``Displays'' are stereotyped sequences of movements that are key to communication between animals of the same species. In penguins, these behaviors are widespread and are used for a variety of purposes including mate choice and pair bonding.
These displays are accompanied by vocalizations that are known to be individually distinctive and to allow both mate and chick recognition in most species \cite{jouventin2017penguins}.

In this paper, we choose to study the ecstatic display (ED), one of the most common and recognizable displays in \emph{Pygoscelid} penguins.
During the ecstatic display penguins stand fully erect on their nests with their stretched neck and bill pointing up vertically.
They move their outstretched flippers back and forth in fast beats while they emit very loud rasps that make their chest vibrate synchronously \cite{jouventin2017penguins,williams95Peng}.
These rasps are normally emitted in pairs of syllables made up of a short inhale followed by a long and loud exhale \cite{bustamante1996vocalizations}.
Here we study this behavior in Chinstrap penguins (\emph{Pygoscelis antarctica}) for which there is an almost complete lack of information regarding their displays \cite{williams95Peng}.

Studies in the two closest species (\emph{Pygoscelis adeliae} and \emph{Pygoscelis papua}) suggest the ecstatic display could be an ``honest display'' intended for males to communicate body conditions to females and/or defend the nesting area from nearby males without a fight \cite{lynch2017variation,Marks2010ecstatic}.
This hypothesis arises because, in those two species, ED occurs only in males at the beginning of the season \cite{williams95Peng}, when they claim a nest and compete to attract a partner.
In chinstrap penguins, however, this behavior happens throughout the season and is displayed by both sexes (as we see through our event camera recordings).
We understand that this behavior could serve a different communication purpose in this species and we want to explore whether it indeed mediates similarly important functions for pair formation and colony structuring as in the other two species.
To find out, first, we must understand when, how, and how long this behavior occurs before drawing relationships to other factors like sex, breeding stage, at-sea behavior, or environmental factors.
Behavioral monitoring like this is proving increasingly important to anticipate changes in breeding habits before a population decline occurs \cite{cerini2023predictive}.
With this work, we aim to open the door for other researchers to use event cameras and TAD for ``large-scale'' automatic detection of behaviors in preventive monitoring. 

\subsection{Results per nest}
\label{sec:suppl:results_per_nest}
\begin{table*}[t]
\centering
\adjustbox{max width=\textwidth}{%
\setlength{\tabcolsep}{4pt}
\begin{tabular}{l*{16}{S}}

\toprule
mAP@IoU                  & \text{N01}  & \text{N02}  & \text{N03} & \text{N04} & \text{N05} & \text{N06} & \text{N07} & \text{N08} & \text{N09} & \text{N10} & \text{N11} & \text{N12} & \text{N13} & \text{N14} & \text{N15} & \text{N16} \\ 
\midrule 
0.1 &  0.528110 & 0.452381 & 0.655123 & 1.000000 & 0.673810 & 0.666667 & 0.854167 & 0.760490 & 0.757576 & 0.976190 & 0.723939 & 0.864073 &0.547619 & 0.719298 & 0.666170 & 0.693846 \\
0.3 &  0.401316 & 0.452381 & 0.655123 & 1.000000 & 0.673810 & 0.666667 & 0.854167 & 0.693182 & 0.725108 & 0.976190 & 0.723939 & 0.864073 & 0.547619 & 0.719298 & 0.647115 & 0.673333 \\
0.5 &  0.297619 & 0.333333 & 0.620622 & 1.000000 & 0.671958 & 0.444444 & 0.583333 & 0.693182 & 0.725108 & 0.976190 & 0.707778 & 0.789957 & 0.547619 & 0.719298 & 0.529718 & 0.589175 \\
0.7 &  0.269737 & 0.027778 & 0.249311 & 0.666667 & 0.671958 & 0.444444 & 0.350000 & 0.469444 & 0.655254 & 0.904762 & 0.430039 & 0.591818 & 0.547619 & 0.719298 & 0.375453 & 0.402381 \\
\textbf{Average}  & 0.374195 & 0.316468 & 0.545045 & 0.916667 & 0.672884 & 0.555556 & 0.660417 & 0.654074 & 0.715762 & 0.958333 & 0.646424 & 0.777481 & 0.547619 & 0.719298 & 0.554614 & 0.589684 \\ [2ex]
\midrule 
\textbf{\# ED}  &    8.000000 &  6.0 &  11.000000 &  3.000000 &  9.000000 &  3.000000 &  4.000000 &  8.000000 &  11.000000 &  6.000000 &  15.000000 &  13.000000 &  7.000000 &  6.000000 &  21.000000 &  10.000000  \\
\bottomrule
\end{tabular}

}
\caption{Results per nest, which show the variation with respect to the data.
}
\label{tab:exp:atsn_per_roi}
\end{table*}
\begin{figure}[t]
    \centering
    {\includegraphics[trim={0cm 0.5cm 0cm 1.3cm},clip,width=0.8\linewidth]{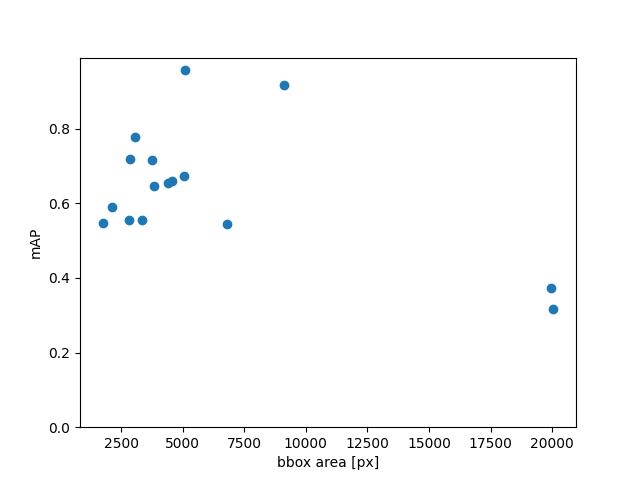}}
    \caption{Visualization showing the relation between bounding box size and mAP.
    } 
    \label{fig:supp:bbox_vs_map}
\end{figure}

The best results of our method per penguin nest are shown in \cref{tab:exp:atsn_per_roi} and visualized in \cref{fig:supp:bbox_vs_map}. 
While it is difficult to extract trends from different nests, the results show a large variation in the data.
It furthermore hints at the importance of good camera positioning during data acquisition.
An elevated camera position, which allows clear separation of the nests, aids in the accurate positioning of the bounding boxes.
This is a lesson learned for future data acquisition campaigns.
We expect to be able to improve the usability of the proposed method by using more cameras and recording fewer nests per camera.

\subsection{Runtime, Computational Effort}
\label{sec:suppl:runtime}
\begin{table}[t!]
\centering
\adjustbox{max width=\columnwidth}{%
\begin{tabular}{lr*{1}{S[detect-weight,detect-mode]}} %
\toprule
Proposal Method           & \# proposals  & \text{AR, Top 50} \\ 
\midrule 
Sliding Window            & 12820320  & 0.078014            \\
Watershed                 & 352       & 0.269504            \\
event TAG \cite{Chen22toc}& 13117     & 0.494090            \\
reTAG (Ours)              & 30527     & \bnum{0.661939}     \\
\bottomrule
\end{tabular}
}
\caption{Number of generated proposals for the test set and the average recall (AR) for different methods.}
\label{tab:suppl:num_proposals}
\end{table}

The number of generated proposals for the test set per method is reported in \cref{tab:suppl:num_proposals}.
We can see the advantage of the TAG-based methods compared to the sliding window approach regarding the number of proposals.
The watershed algorithm is too simple and does not produce a sufficient amount of proposals.
Comparing the two TAG-based algorithms shows the trade-off between recall and computational effort.
While our reTAG has significantly increased recall, it also generated five times more proposals.
Overall, our reTAG is lightweight compared to the sliding window approach and has $8\times$ higher average recall.

\textbf{Inference time}: The ATSN with a ResNet18 backbone has a run time of 2.56ms (Nvidia Tesla V100S) for one example and one forward pass.
Each time interval generated in the first step (proposal) is a sample for training and validation of the second step (classifier).

\textbf{Training time}: For the training set the reTAG algorithm outputs 189519 proposals.
To maintain a manageable balance between foreground (ED) and background, the negative samples in the training set are sub-sampled by a factor of 10, leading to a training set with 2093 positive and 18859 negative samples. 
We train for 10 epochs, resulting in a training time for the classifier of approximately 30 minutes on an Nvidia Tesla V100S GPU.

\subsection{Dataset Details}
\label{sec:suppl:dataset_details}

In total, we collected around 238 hours of unfiltered data from the Vapour Col penguin colony in Deception Island, in the form of ROSBag files.
The ROSBag files were then post-processed using a hot pixel filter \cite{Scheerlink19github}, which discards data from faulty pixels that trigger events at a high rate.
Among the recorded data, we selected 24 ten-minute-long sequences for annotation of ecstatic display (ED) behavior.
The selected sequences include diverse scenarios from different dates and hours of the day, to account for various illumination and weather conditions.

Incidentally, our setup was deployed next to a foraging camera which takes a snapshot of the penguin colony every 1 minute.
During the night, the foraging camera uses flashes of infrared (IR) light to acquire images in low light.
Since the event camera sensor is also sensitive to IR light, the aforementioned flashes produce a flurry of events throughout the scene once per minute.
Consequently, events generated by IR flashes were filtered out from the night sequences before applying our proposed TAD algorithm.

Detailed information on every ten-minute sequence in the annotated dataset can be found in \cref{tab:suppl:datasplit}. 
The table furthermore indicates the data split (training, validation, testing) we used in all experiments.
The test split reflects the proportion of precipitation in Antarctica during the breeding season.

\begin{table}[t!]
\centering
\adjustbox{max width=\columnwidth}{%
\setlength{\tabcolsep}{4pt}
\begin{tabular}{l*{5}{S}}
\toprule
Day    &   \text{time}    &   \text{split} &   \text{night}   &   \text{precipitation}  &  \text{\#ED} \\
 \midrule 
Jan 5th &  \text{17:00}  & \text{train}    &   \xmark         &    \xmark      &   \text{2}     \\ 
Jan 6th &  \text{19:00}  & \text{train}    &   \xmark         &    \xmark      &  \text{11}      \\ 
Jan 7th &  \text{05:00}  & \text{train}    &   \xmark         &    \cmark      &   \text{70}     \\ 
Jan 7th &  \text{08:00}  & \text{train}    &   \xmark         &    \xmark      &  \text{3}      \\ 
Jan 9th &  \text{20:04}  & \text{train}    &   \xmark         &    \xmark      &  \text{28}      \\ 
Jan 11th &  \text{21:06}  & \text{train}    &   \xmark         &    \xmark     &   \text{9}      \\
Jan 12th &  \text{03:36}  & \text{train}    &   \cmark         &    \cmark     &   \text{54}      \\
Jan 12th &  \text{03:56}  & \text{train}    &   \xmark         &    \cmark     &   \text{23}      \\ 
Jan 12th &  \text{08:56}  & \text{train}    &   \xmark         &    \cmark     &   \text{0}      \\
Jan 12th &  \text{12:56}  & \text{train}    &   \xmark         &    \xmark     &  \text{0}       \\
Jan 12th &  \text{17:26}  & \text{train}    &   \xmark         &    \xmark     &   \text{58}      \\
Jan 13th &  \text{00:00}  & \text{train}    &   \cmark         &    \xmark     &  \text{92}       \\
Jan 13th &  \text{10:59}  & \text{train}    &   \xmark         &    \xmark     &   \text{3}      \\
Jan 13th &  \text{14:59}  & \text{train}    &   \xmark         &    \cmark     &   \text{11}      \\
Jan 14th &  \text{23:58}  & \text{train}    &   \cmark         &    \xmark     &   \text{1}      \\
Jan 15th &  \text{13:58}  & \text{train}    &   \xmark         &    \xmark     &   \text{0}      \\
Jan 18th &  \text{02:56}  & \text{train}    &   \cmark         &    \xmark     &  \text{0}       \\
\midrule 
Jan 7th &  \text{02:00}  & \text{validation}    &   \cmark         &    \xmark   &  \text{20}     \\ 
Jan 17th &  \text{15:56}  & \text{validation}    &   \xmark         &    \xmark  &  \text{29}     \\ 
\midrule
Jan 6th &  \text{01:00}  & \text{test}    &   \cmark         &    \xmark       &   \text{53}    \\ 
Jan 13th &  \text{09:59}  & \text{test}    &   \xmark         &    \xmark      &   \text{47}     \\ 
Jan 14th &  \text{21:58}  & \text{test}    &   \xmark         &    \xmark      &   \text{8}     \\ 
Jan 15th &  \text{05:58}  & \text{test}    &   \xmark         &    \xmark      &   \text{18}     \\ 
Jan 15th &  \text{11:48}  & \text{test}    &   \xmark         &    \cmark      &   \text{25}     \\ 
\bottomrule
\end{tabular}
}
\caption{An overview of all ten-minute sequences in the annotated dataset.}
\label{tab:suppl:datasplit}
\end{table}

\subsection{Additional Sensitivity Studies}
\label{sec:supp:sensitivity_backbone}

\Cref{tab:exp:whole-pipeline:mobilenet} reports the results using a MobileNetV3 backbone instead of ResNet18.
The figures are similar with both backbone variants and event representations (only a 4\% performance drop on average), supporting the robustness of our method to different design choices.

Similarly, \cref{tab:supp:sensitivity_deltat} shows results concerning different values of the accumulation time $\Delta t$ for histograms and decay $\tau$ for time maps.

\begin{table}[ht]
\centering
\adjustbox{max width=\columnwidth}{%
\setlength{\tabcolsep}{4pt}
\begin{tabular}{ll*{5}{S[detect-weight,detect-mode]}} %
\toprule
Backbone & Event repres. & \text{0.1} & \text{0.3} & \text{0.5} & \text{0.7} & \text{Average} \\
\midrule 
ResNet18 (Tab. 3) & Time-map &  \bnum{0.6595} & \bnum{0.6379} & \bnum{0.581} & \bnum{0.4251}  & \bnum{0.5759} \\
MobileNetV3-Large & Time-map  & 0.6232 & 0.5914 & 0.5259 & 0.3648 & 0.5263 \\
MobileNetV3-Large & Histogram & 0.5737 & 0.535 & 0.493 & 0.3646 & 0.4916 \\
MobileNetV3-Small & Time-map  &  0.6021 & 0.5607 & 0.4952 & 0.3479 & 0.5015 \\
MobileNetV3-Small & Histogram &  0.5046 & 0.4835 & 0.4494 & 0.3199 & 0.4393 \\
\bottomrule
\end{tabular}
}
\caption{Sensitivity of the system with respect to the backbone (ResNet or MobileNet) 
and input represesntation.
Mean Average Precision at several IoU levels (mAP@IoU).}
\label{tab:exp:whole-pipeline:mobilenet}
\end{table}

\begin{table}[th]
\centering
\adjustbox{max width=\linewidth}{
\begin{tabular}{lcccrccc}
\toprule
  & \multicolumn{3}{c}{Histogram $\Delta t$} & & \multicolumn{3}{c}{Time map: Decay $\tau$ [s]} \\ 
  \cmidrule(l{1mm}r{1mm}){2-4} \cmidrule(l{1mm}r{1mm}){6-8}
 $\Delta t$ [s] & 0.3  & 1    &   3   & $\tau$ [s] & 0.02  & 0.2  & 2  \\ 
 mAP            & 0.55 & 0.56 &  0.52 & mAP & 0.45  & 0.58 & 0.51 \\ 
\bottomrule
\end{tabular}
}
\caption{Sensitivity of the system concerning parameters of the frame representation.
Mean average precision at several IoU levels (mAP@IoU).}
\label{tab:supp:sensitivity_deltat}
\end{table}

\subsection{Naive Baseline Classifier}
\label{sec:suppl:naive_baseline}

There is a high class imbalance in the proposals. 
Randomly guessing leads to a high number of false positives, and consequently a low precision.
To confirm this, we implemented a random classifier (akin to flipping a coin), which accepts a proposal with a 50\% change, setting the score to 1 for mAP calculation.
This solution achieves 0.035\% mAP (Avg.).

\fi

{\small
\bibliographystyle{ieeetr_fullname} %
%\bibliography{all,egbib}

}

\end{document}